\documentclass[letterpaper, 10 pt, conference]{ieeeconf}  

\IEEEoverridecommandlockouts                              

\usepackage{amsmath} 
\usepackage{amssymb}  
\usepackage{amsmath}
\usepackage{hyperref}
\usepackage{amsmath}
\usepackage{graphicx}
\usepackage{tikz}
\usepackage{bm}
\usepackage{multicol}
\usepackage{listings}
\usepackage[skip=2pt,font=scriptsize]{caption}
\usetikzlibrary{fit, automata, positioning}
\title{\LARGE \bf
Closed-Loop Control of a Delta-Wing Unmanned Aerial-Aquatic Vehicle
}

\author{Joseph Moore$^{1}$ 
\thanks{*This work was supported by the Johns Hopkins University Applied Physics Lab Internal Research and Development Program}
\thanks{$^{1}$Joseph Moore is is a Research Scientist in the Intelligent Systems Group, Johns Hopkins University Applied Physics Lab, Laurel, MD, 20723
        {\tt\small joseph.moore@jhuapl.edu}}%
}

\newcommand{\matr}[1]{\mathbf{#1}} 
\newcommand{\matrs}[1]{\bm{\mathcal{#1}}}
\newcommand{\vect}[1]{\mathbf{#1}} 
\newcommand{\vectg}[1]{\boldsymbol{#1}}

\newcommand{\air}{\mathrm{a}}
\newcommand{\water}{\mathrm{w}}
\newcommand{\bx}{\mathbf{x}}
\newcommand{\bu}{\mathbf{u}}
\newcommand{\bff}{\mathbf{f}}
\newcommand{\bR}{\mathbf{R}}

\newcommand{\bex}{{\bf e}_{x}}

\newcommand{\bezr}{{\bf e}_{z_r}}
\newcommand{\bexd}{{\bf e}_{x_\delta}}

\newcommand{\bpi}{{\boldsymbol \pi}}

\begin{document}

\maketitle
\thispagestyle{empty}
\pagestyle{empty}

\begin{abstract}

We present a closed-loop control strategy for a delta-wing unmanned aerial aquatic-vehicle (UAAV) that enables autonomous swim, fly, and water-to-air transition. Our control system consists of a hybrid state estimator and a closed-loop feedback policy which is capable of trajectory following through the water, air and transition domains. To test our estimator and control approach in hardware, we instrument the vehicle presented in \cite{moore2018design} with a minimalistic set of commercial off-the-shelf sensors. Finally, we demonstrate a successful autonomous water-to-air transition with our prototype UAAV system and discuss the implications of these results with regards to robustness.


\end{abstract}

\section{INTRODUCTION}
Unmanned aerial-aquatic vehicles (UAAVs) have the potential to dramatically improve remote access of underwater environments. In particular, \emph{fixed-wing} UAAVs offer a promising means of enabling efficient locomotion in both aerial and aquatic domains through the use of a lifting surface. In a previous paper \cite{moore2018design}, we presented a propeller-driven delta-wing unmanned aerial-aquatic vehicle design and demonstrated that, given the right parameter values, successful water-exit could be achieved for a given maximum thrust available in water and air. However, unlike other approaches which rely on novel propulsion mechanisms to achieve water-exit \cite{izraelevitz2015novel,siddall2014launching,maia2015demonstration}, our design relies on the availability of closed-loop feedback control to enable multi-domain locomotion. In our prior paper, we asserted that a software-only water-exit solution had the potential to greatly reduce the cost and mechanical complexity of hybrid aerial-aquatic vehicles. In this paper, we present a feedback control approach for enabling a water-to-air transition. To limit the scope of this work, we focus strictly on the water-exit problem.

Our approach consists of a nominal trajectory generated using the vehicle's hybrid dynamics and an optimal time-varying feedback controller for trajectory following. A hybrid state estimator is used to observe the necessary vehicle states and hybrid modes. We first demonstrate our control approach in simulation. We then instrument our vehicle and demonstrate successful state estimation across the domain transition. Finally, we demonstrate closed-loop control in hardware and show a successful autonomous water-exit.

 \begin{figure}
\centering
 \includegraphics[width=2.75in,trim={0mm 10mm 0mm 10mm},clip]{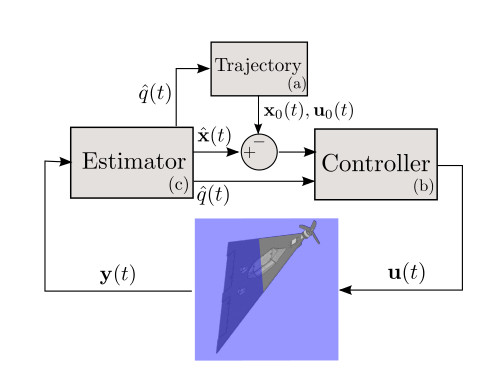}
  \caption{A depiction of the closed-loop control system used to execute a water-exit maneuver. (a) provides a feasible multi-domain water-exit trajectory, (b) represents the hybrid control system, and (c) is the state estimator for estimating the continuous system state and the mode.}
  \label{fig:control_system}
\end{figure}

\section{RELATED WORK}
Due to advances in technology, especially in the remote-controlled aircraft domain, several unmanned aerial-aquatic vehicles have emerged in the past few years \cite{yang2015survey}. These vehicles, which span the air and water domains, have been the subject of both design studies and hardware demonstrations. Many design strategies have focused on novel propulsion mechanisms. In \cite{maia2015demonstration}, the authors develop a quadcopter which can both swim and fly, and is able to transition between the two domains via a novel propeller design. In \cite{siddall2015water,siddall2014launching,siddall2016fast}, the authors develop a fixed-wing UAAV which uses a water-bottle rocket-like propulsion mechanism to exit the water. Some of the same authors present a novel gearbox design to enable multi-domain locomotion with a single propeller in \cite{tan2017efficient}. In \cite{izraelevitz2015novel}, the authors propose a flapping multi-domain wing design and discuss its implications for multi-domain locomotion. Researchers have also engaged in structural analysis of these aerial-aquatic systems. For instance, in \cite{yang2013computational}, the authors present a computational analysis of a fixed-wing UAAV impacting the water during water-entry.

Few of the approaches mentioned above consider the closed-loop control or estimation strategies necessary for enabling multi-domain locomotion. In \cite{drews2014hybrid} and \cite{neto2015attitude} the authors present modeling, simulation and control strategies for a multi-domain quadcopter. Their approach focuses on applying robust control techniques to develop a globally stable switching attitude controller which relies on two different linear models. In \cite{ravell2018modeling}, the authors explore hybrid control for the quadrotor UAAV experimentally. 


Our delta-wing UAAV design favors a simplified mechanical and propulsion design and the use of commercial off-the-shelf components to facilitate domain transitions. Because automatic control is key to enabling a water-to-air transition with our system, a strategy for controlling the vehicle across the water-air interface must be explored. In this work, we present such a feedback control system for a fixed-wing UAAV and experimentally demonstrate a successful autonomous water-exit. Our control approach reasons about the nonlinear dynamics of the vehicle in both domains and attempts to control both attitude, altitude and velocity.

The recent work most similar to our approach are \cite{peloquin2017design} and \cite{weisler2017testing}. Both explore high thrust-to-weight ratio tail-sitter fixed-wing UAV designs. The vehicle presented in \cite{peloquin2017design} is distinct in that it is not submersible, but uses a novel passive mechanism to facilitate rapid take-off from the water's surface. The vehicle in \cite{weisler2017testing} also possesses some novel design attributes--- notably passive water draining from inside the wing's cavity. While this fixed-wing vehicle is submersible, it differs from our system in some important ways.


First, our vehicle possesses a delta-wing planform, which results in different aero- and hydrodynamic characteristics, especially during the transition domain. Second, to take advantage of the ``buoyancy assist'' described in \cite{moore2018design}, unlike \cite{weisler2017testing}, our vehicle has the center of buoyancy at the rear, which requires active stabilization of the vehicle while underwater to achieve successful water-exit. Third, the water-to-air transition in \cite{weisler2017testing} occurs over a longer time-period ($\approx$ 2s) and starts vertically from rest; our vehicle's water-to-air transition occurs in less than 0.5s from an underwater cruise velocity. We believe that this dynamic transition coupled with the delta-wing design has the potential to dramatically reduce energy consumption during water egress. However, the dynamic nature of this transition and the passive underwater instability of our system requires a control system that can explicitly reason about the vehicle's hybrid dynamics.

In this paper, we focus on such a control system. We limit our attention to the water-exit problem and develop a controller which both estimates the hybrid modes and reasons about the physics of the vehicle in these different domains to achieve dynamic water-to-air transition with a fixed-wing UAAV.



\section{VEHICLE MODEL}
In our previous work \cite{moore2018design}, a multi-domain model of the delta-wing unmanned aerial-aquatic vehicle was presented. In that work, the model was used to compute a water-exit trajectory to validate our design. In this paper, the model serves as the basis for our control design, enabling generation of the nominal trajectory as well as computation of the feedback gains. Here, we briefly revisit the model presented in \cite{moore2018design}, and show how it is used for control design. 

We define our state as
$\vect{x} = \{ r_x, r_y, r_z, \phi, \theta, \psi, \delta_{1}, \delta_{2}, v_{x}, v_{y}, v_{z}, \omega_x, \omega_y, \omega_z\}$.
Here $\vect{r} =\begin{bmatrix} r_x,r_y,r_z\end{bmatrix}^T$ represents the position of the center of mass in the world frame $O_{x_r y_r z_r}$, $\vectg{\theta}= \begin{bmatrix} \phi, \theta, \psi\end{bmatrix}^T$ represents the set of $z$-$y$-$x$ Euler angles, $\vectg{\delta} = \begin{bmatrix} \delta_1, \delta_2 \end{bmatrix}^T$ are control surface deflections due to the right and left elevons, $\vect{v} = \begin{bmatrix} v_x, v_y, v_z\end{bmatrix}^T$ is the velocity of the center of mass in the \emph{body} fixed frame $O_{xyz}$, $\vectg{\omega} = \begin{bmatrix}\omega_x, \omega_y, \omega_z\end{bmatrix}^T$ represents the angular velocity of the body the body-fixed frame. We can then write $\vect{x} = \{\vect{r}^T, \vectg{\theta}^T, \vectg{\delta}^T, \vectg{v}^T, \vectg{\omega}^T\}^T$. The control input is $\vect{u}= \begin{bmatrix}\vect{u}_{cs}^T, \vectg{\delta}_t\end{bmatrix}^T$, where $\vect{u}_{cs}^T$ contains the control surface velocities as $\begin{bmatrix}\dot{\delta}_1, \dot{\delta}_2 \end{bmatrix}$ and $\delta_t$ is the thrust of the propeller.

\begin{figure}[t]
\centering
\begin{tikzpicture}[shorten >=1pt,node distance=2.5cm,on grid,auto, scale = 0.5, every node/.style={scale=0.5}] 
   \node[state] (q_3) [align=center,label={[align=center]left:\includegraphics[width=.15\textwidth]{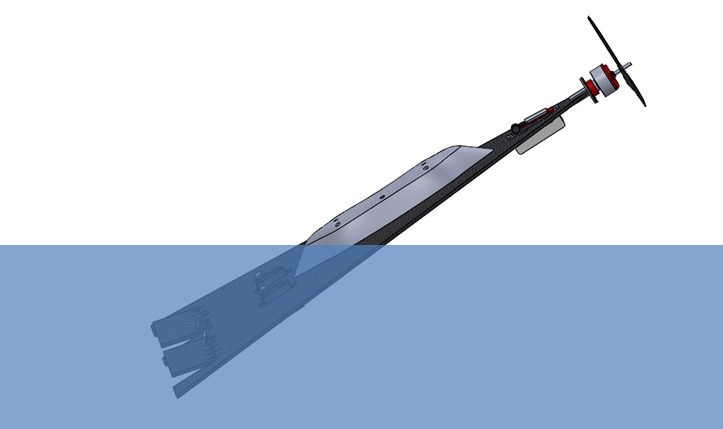}\\ \textbf{transition}$_A$}]   {$q=q_1$\\$\dot{\bx}=\bff(\bx,\bu,q)$\\$\Psi_1(\bx,\bu)>0$\\$\Psi_2(\bx,\bu)<0$};
   \node[state] (q_0) [align=center, above right=of q_3,label={[align=center]above:\includegraphics[width=.15\textwidth]{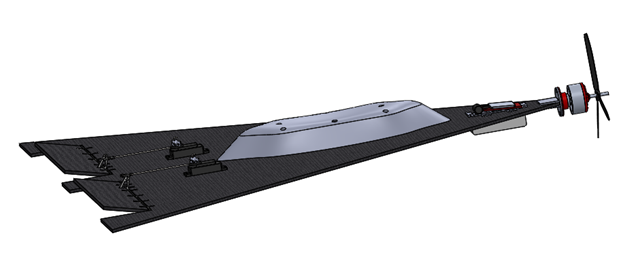}\\ \textbf{air}}] {$q=q_2$\\$\dot{\bx}=\bff(\bx,\bu,q)$\\$\Psi_1(\bx,\bu)>0$\\$\Psi_2(\bx,\bu)>0$}; 
   \node[state] (q_1) [align=center,below right=of q_0,label={[align=center]right:\includegraphics[width=.15\textwidth,trim={0 0 0 0}]{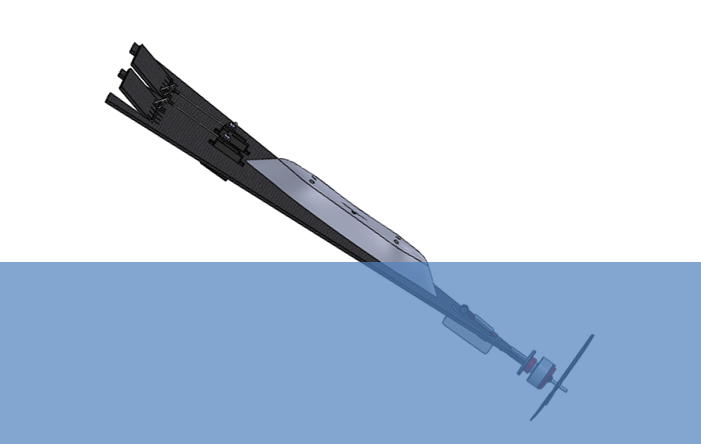}\\ \textbf{transition}$_B$}] {$q=q_3$\\$\dot{\bx}=\bff(\bx,\bu,q)$\\$\Psi_1(\bx,\bu)<0$\\$\Psi_2(\bx,\bu)>0$}; 
   \node[state] (q_2) [align=center,below right=of q_3,label={[align=center]below:\includegraphics[width=.15\textwidth]{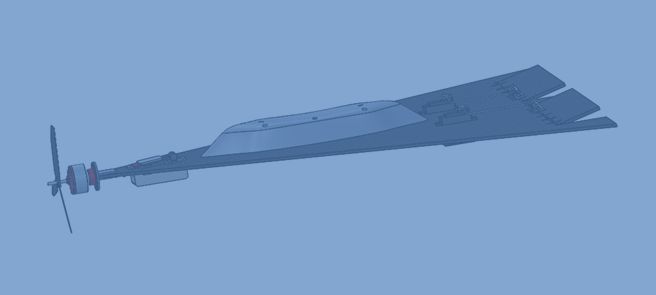}\\ \textbf{water}}] {$q=q_0$\\$\dot{\bx}=\bff(\bx,\bu,q)$\\$\Psi_1(\bx,\bu)>0$\\$\Psi_2(\bx,\bu)<0$};
    \path[->] 
    (q_3) edge  [align=center, bend left, above] node {} (q_0)
    (q_0) edge  [align=center, bend left, above] node {} (q_3)
    
    (q_0) edge  [align=center, bend left, above] node {} (q_1)
    (q_1) edge  [align=center, bend left, above] node {} (q_0)
    
    (q_1) edge  [align=center, bend left, above] node {} (q_2)
    (q_2) edge  [align=center, bend left, above] node {} (q_1)
    
    (q_3) edge  [align=center, bend left, above] node {} (q_2)
    (q_2) edge  [align=center, bend left, above] node {} (q_3)
          ;
\end{tikzpicture}
\caption{Depiction of the hybrid dynamical system described by the physics model.}
\label{fig:hybrid2}
\end{figure}

The equations of motion then become
\begin{align}
\dot{\vect{r}} &= \bR_b^r \vect{v}\nonumber\\
\dot{\vectg{\theta}} &= {\bR}_{\vectg{\omega}}^{-1}{\boldsymbol \omega}\nonumber\\
\dot{\vectg{\delta}} &= \vect{u}_{cs}\nonumber\\
\dot{\vectg{\chi}} &= (\matr{M}+\matr{M}_a)^{-1}\big(\vectg{f}-\matrs{S}({\vectg{\omega}})(\matr{M}+\matr{M}_a)\vectg{\chi}\\&\quad-\matrs{S}({\vect{v}})\matr{M}_a\vectg{\chi}\big)\nonumber
\end{align}
where
\begin{align}
\vectg{\chi}=\begin{bmatrix}\vect{v}, \vectg{\omega}\end{bmatrix}^T, \quad
\matr{M}=\begin{bmatrix}m\matr{I}  & \bm{0}\\ \bm{0} & \matr{J}\end{bmatrix},\quad
\vectg{f}=\begin{bmatrix}\vect{f}, \vect{m}\end{bmatrix}^T\\
\matrs{S}({\vectg{\omega}})=
\begin{bmatrix}\matr{S}(\vectg{\omega})  & \bm{0}\\ \bm{0} & \matr{S}(\vectg{\omega}) \end{bmatrix},\quad
\matrs{S}({\vect{v}})=
\begin{bmatrix}\bm{0}  & \bm{0}\\ \matr{S}({\vect{v}}) & \bm{0} \end{bmatrix}.
\end{align}
$\matr{M}_a$ is the ``added mass'' matrix, $m$ is the vehicle mass, $\matr{J}$ is the vehicle's inertia tensor with respect to the center of mass, $\vect{f}$ is the total force (excluding the forces due to added mass) applied to the vehicle in body-fixed coordinates, $\vect{m}$ are the moments applied about the vehicle's center of mass in body-fixed coordinates, $\matr{S}(\vectg{\omega})=\vectg{\omega}_{\times}$ and $\matr{S}(\vect{v})=\vect{v}_{\times}$.
$\bR_b^r$ denotes the rotation from the body-fixed frame to the world frame, and ${\bR}_{\vectg{\omega}}$ is the rotation which maps the euler angle rates to an angular velocity in the body-fixed frame. The forces $\bff$ and moments $\vect{m}$, defined explicitly in \cite{moore2018design}, are dependent on the density of the fluid surrounding the control and lifting forces. To capture this density change, we model our vehicle as a hybrid system (see Figure \ref{fig:hybrid2}).

This hybrid system can be modeled compactly as
\begin{align}
\dot{\bx}=\bff(\bx,\bu,q)\nonumber\\
q = \matr{\Phi}(\bx^{-},q^{-})\nonumber\\
\bx^{+}= \Delta(\bx^{-},q^{-})
\end{align}
where $q$ is the mode of the system, $q^{-}$ is the mode just prior to the transition, $\bx^{-}$ is the state prior to a transition and $\bx^{+}$ represents the state after a mode transition occurs. $\rho_A$, $\rho_F$, and $\rho_{\delta_i}$ represent the density of the fluid surrounding the fore, aft, and the $i^{th}$ control surface respectively. They can be defined using the hybrid mode $q$ as

\begin{align}
\vectg{\rho}^T=\begin{bmatrix}\rho_A\\ \rho_F\\ \rho_{\delta} \end{bmatrix}^T=
\begin{cases}
\vspace{1mm} 
\begin{bmatrix}\rho_\air& \rho_\air &\rho_\air\end{bmatrix} & q=q_0\\
\vspace{1mm}
\begin{bmatrix}\rho_\water& \rho_\air &\rho_\air\end{bmatrix} & q=q_1\\
\vspace{1mm}
\begin{bmatrix}\rho_\water& \rho_\water &\rho_\water\end{bmatrix} & q=q_2\\
\vspace{1mm}
\begin{bmatrix}\rho_\air& \rho_\water &\rho_\water\end{bmatrix} & q=q_3\\
\end{cases}
\end{align}
where $\rho_\water =1000 \frac{kg}{m^3}$, $\rho_\air = 1.22 \frac{kg}{m^3}$. $\rho_{\delta_i} =\rho_{\delta}$ for the delta-wing executing a planarized transition maneuver. The mode transition function is given as


\begin{align}
q =\Phi(\bx^{-},q^{-})=
\begin{cases} 
q_0 & q_1, \Psi_{1,0}\geq 0\\
q_0 & q_3, \Psi_{3,0}\geq 0\\
q_1 & q_0, \Psi_{0,1}\geq 0\\
q_1 & q_2, \Psi_{2,1}\geq 0\\
q_2 & q_1, \Psi_{1,2}\geq 0\\
q_2 & q_3, \Psi_{3,2}\geq 0\\
q_3 & q_2, \Psi_{2,3}\geq 0\\
q_3 & q_0, \Psi_{0,3}\geq 0\\
\end{cases}
\end{align}
where
\begin{align}
\bar{\Psi}_0 &= \begin{bmatrix}\Psi_{0,1}\\ \Psi_{0,3}\end{bmatrix} = \begin{bmatrix}\Psi_1\\ -\Psi_2\end{bmatrix}, \quad \bar{\Psi}_1 = \begin{bmatrix}\Psi_{1,0}\\\Psi_{1,2}\end{bmatrix} = \begin{bmatrix}-\Psi_1\\ \Psi_2\end{bmatrix}\\
\bar{\Psi}_2 &= \begin{bmatrix}\Psi_{2,3}\\ \Psi_{2,1}\end{bmatrix} = \begin{bmatrix}-\Psi_1\\-\Psi_2\end{bmatrix}, 
\quad \bar{\Psi}_3 = \begin{bmatrix}\Psi_{3,2}\\ \Psi_{3,0}\end{bmatrix} = \begin{bmatrix}\Psi_1\\ \Psi_2\end{bmatrix}\nonumber
\end{align}
and
\begin{align*}
\Psi_1(\bx) = \bezr^T(\vect{r} + \bR_b^r(L\bex-l_{cg}\bex))\\
\Psi_2(\bx) = \bezr^T(\vect{r} + \bR_b^r\vect{r}_h + \bR_b^r\bR_{\delta}(-l_{\delta}\bexd)).
\end{align*}
Here $\bezr$ is the unit vector in the $z_r$ direction, $\bex$ is the unit vector in the $x$ direction, $L$ is the length of the wing, $l_{cg}$ is the distance from the trailing edge to the center of mass, $\vect{r}_h$ is the displacement from the center of mass to the elevon hinge $\bR_{\delta}$ is the rotation matrix from the control surface frame to the body frame, $l_{\delta}$ is the length from the elevon hinge to the elevon center of pressure, and $\bexd$ is the unit vector aligned with the x-coordinate of the elevon. 
The reset map is simply given as $\bx^{+}= \bx^{-}$
\section{TRAJECTORY OPTIMIZATION}
At the core of our control algorithm is a feasible optimal multi-domain trajectory (see Figure \ref{fig:control_system}). In \cite{moore2018design}, we used a trajectory optimization approach to demonstrate that, given the aforementioned model, a water-exit maneuver was possible. Here, we again generate a water-exit trajectory for a vehicle; however, this time it will be used for feedback and feedfoward control. 

To reduce the number of parameters in the optimization problem, we utilize a planarized model which only considers the vehicle's \emph{longitudinal} dynamics. As before, to design our trajectory, we use a direct formulation known as direct transcription so that hard constraints can be imposed on state. 

We provide our hybrid trajectory optimizer with a hybrid dynamical system (represented as above), an initial condition set, a final condition set, and a mode schedule. The trajectory optimizer is then able to produce a feasible multi-domain trajectory. We formulate the trajectory optimization problem as follows:

\subsection{Problem Formulation}
Assuming a feasible mode schedule, $\xi= \{\xi_0,\xi_1,\xi_2... \}$, we construct the following optimization problem: Let $n_ih$ represent a ``sub'' time horizon for a particular mode in the schedule where $N_j=\sum_{i=0}^{j} n_i$, $M_j=N_j-n_j$, $O=|\xi|$, $N = N_O$.
Our cost-function can be written as
\begin{equation}
\begin{aligned}
& \underset{\vect{x}_k, \vect{u}_k, h_j }{\text{min}}
& & g_f(\vect{x}_{N}) + \sum_{j=0}^{O-1} \sum_{k=M_j}^{N_j-1} g(\vect{x}_{k}, \vect{u}_{k},  h_j) \\
& \text{s.t.}
& & \vect{x}_{k} - \vect{x}_{k+1} + \frac{h_j}{6.0}(\vect{\dot{x}}_{k} + 4 \vect{\dot{x}}_{c,k} + \vect{\dot{x}}_{k+1})= 0,\\
&
& & \forall k \in \{M_j,\ldots N_j-1\} \text{ and } \forall j \in \{ 1,\ldots O-1\}\\
&
& &\bar{\Psi}_{\xi_j}(\vect{x}_{k}) \le 0, \quad \forall j \in \{ 0,\ldots O-1\}\\ 
&
& &\Psi_{\xi_j, \xi_{j+1}}(\vect{x}_{N_j}) = 0, \quad \forall j \in \{ 0,\ldots O-2\}\\
&
& & \vect{x}_f - \vectg{\delta}_f \le \vect{x}_{N} \le \vect{x}_f + \vectg{\delta}_f\\
&
& & \vect{x}_i - \vectg{\delta}_i \le \vect{x}_{N} \le \vect{x}_i + \vectg{\delta}_i\\
&
& & \vect{x}_{min}  \le \vect{x}_{k} \le \vect{x}_{max},~~ \vect{u}_{min}  \le \vect{u}_{k} \le \vect{u}_{max}\\
&
& & h_{min}  \le h_j \le h_{max}
\end{aligned}
\end{equation}

where 
\begin{align}
\vect{\dot{x}}_{k} &= \vect{f}(t, \vect{x}_{k}, \vect{u}_{k}, \xi_j),~~
\vect{\dot{x}}_{k+1} = \vect{f}(t, \vect{x}_{k+1}, \vect{u}_{k+1}, \xi_j)\nonumber\\
\vect{u}_{c,k} &= (\vect{u}_{k} + \vect{u}_{k+1}) / 2\nonumber\\
\vect{x}_{c,k} &= (\vect{x}_{k} + \vect{x}_{k+1}) / 2 + h_j (\vect{\dot{x}}_{k} - \vect{\dot{x}}_{k+1}) / 8\nonumber\\
\vect{\dot{x}}_{c,k} &= \vect{f}(t, \vect{x}_{c,k}, \vect{u}_{c,k}, \xi_j)
\end{align}
The cost function terms can be defined as
\begin{align}
g(\vect{x}_{k}, \vect{u}_{k},  h_j)= \vect{u}_{k}^T\vect{R}\vect{u}_{k}h_j + Dh_j, \quad g(\vect{x}_{N}, \vect{u}_{N}) = {\bf 0}.
\end{align}
This formulation of the optimization problem allows for trajectory segments to be separated into different modes, while maintaining state continuity across modes and minimizing the control (thrust) effort.
\begin{align}
&\vect{x}_i=[-3.5,-1, 0, 0, 0.5, 0, 0]^T \nonumber\\
&\vectg{\delta}_i=[0.5, 0.1, 0.05, 0, 0, 0, 0]^T\nonumber\\
&\vect{x}_f=[0, 1, 0, 0, 10, 0, 0]^T\nonumber\\
&\vect{\delta}_f=[2, 0.5, 0.15, \pi/2, 2, 2, 10]^T\nonumber\\
&\vect{x}_{max}=-\vect{x}_{min}=[10, 10, 10, 10, 10, 10, 10]^T\nonumber\\
&\vect{u}_{max}=[10, 10, 5]^T\quad\vect{u}_{min}=[-10,-10, 0]^T\nonumber \\ 
&D = 1\quad\vect{R}= \matr{I}
\label{problem:traj_opt}
\end{align}
The trajectory optimizer is implemented in C++, and the nonlinear optimization problem is solved using the Sparse Nonlinear OPTimizer (SNOPT) \cite{gill2005snopt}.

\section{FEEDBACK CONTROL}
The trajectory provided by solving \ref{problem:traj_opt} provides the vehicle with a feasible nominal trajectory to follow. However, given model and environmental uncertainty, feedback for trajectory tracking will be necessary to ensure a successful water-to-air transition. In addition, ensuring that the system successfully makes the domain transitions will also be critical. To design this feedback strategy, we assume that we have a mode schedule and a means of detecting the mode transitions. We will attempt to find a set of time-varying gains to stabilize a trajectory, and we will place a time-invariant controller on the guards to make these domain transitions ``attractive''. This feature will be very helpful in making sure that the mode changes occur successfully. A diagram of our feedback control approach can be seen in Figure \ref{fig:hybridcontroller}.

Our controller, $\bpi(\tau, \bx,q)$ can be written as:  

\begin{figure}[t]
\centering
\begin{tikzpicture}[shorten >=1pt,node distance=2.5cm,on grid,auto, scale = 0.5, every node/.style={scale=0.5}] 
   \node[state] (q_0) [align=center,label={[align=center]}]   {$m=m_{j}$\\$q=q_{j}$\\$\bpi=\bpi_{\tau}(\tau,\bx,q)$\\$\dot{\tau}=1$};
   \node[state] (q_1) [align=center, above right=of q_0,label={[align=center]}] {$m=m_{T,j}$\\$q=q_{j}$\\$\bpi=\bpi_{T}(\bx,q)$\\$\dot{\tau}=0$}; 
   \node[state] (q_2) [align=center,below right=of q_1,label={[align=center]}] {$m=m_{k}$\\$q=q_k$\\$\bpi=\bpi_{\tau}(\tau, \bx,q)$\\$\dot{\tau}=1$}; 
   \node[] (q_3) [align=center, left=of q_0,label={[align=center]}] {};
   \node[] (q_4) [align=center,above left=of q_0,label={[align=center]}] {};
   \node[] (q_5) [align=center, right=of q_2,label={[align=center]}] {};
   \node[] (q_6) [align=center,above right=of q_2,label={[align=center]}] {};
    \path[->] 
    (q_0) edge  [align=center, bend left, below, pos = 0.25] node {$\qquad \qquad \tau\geq T(q_j)$\\} (q_1)
    (q_1) edge  [align=center, bend left, above, pos = 0.25] node {$\quad\qquad\Psi_{j,k}(x^{-})\geq 0$\\} (q_2)
    (q_1) edge  [align=center, bend left, below, pos = 0.8] node {$\tau:=0\quad\qquad$} (q_2)
    (q_0) edge  [align=center, bend left, below, pos = 0.25] node {\\ \\$\Psi_{j,k}(x^{-})\geq 0$} (q_2)
    (q_0) edge  [align=center, bend left, below, pos = 0.85] node {\\ $\tau:=0$} (q_2)

    (q_3) edge  [align=center, bend left, below, pos = 0.25] node {\\$\qquad \qquad\Psi_{i,j}(x^{-})\geq 0$} (q_0)
    (q_4) edge  [align=center, bend left, above, pos= 0.25] node {$\Psi_{i,j}(x^{-})\geq 0$} (q_0)
    (q_3) edge  [align=center, bend left, above, pos=0.85] node {$\tau:=0$} (q_0)
    (q_4) edge  [align=center, bend left, above,pos=0.95] node {$\tau:=0\qquad \qquad$} (q_0)
    (q_2) edge  [align=center, bend left, below, pos=0.3] node {\\ \\$\Psi_{k,l}(x^{-})\geq 0$} (q_5)
    (q_2) edge  [align=center, bend left, below, pos=0.2] node {$\quad \qquad \qquad\tau>T(q_k)$\\} (q_6)
          ;
\end{tikzpicture}
\caption{Diagram of a hybrid control system for a generic mode schedule of the dynamical system $\xi=\{\xi_i,\xi_j,\xi_k,\xi_l\}$. The nodes in the graph represent the hybrid mode state, $m$, of the closed-loop dynamical system where the time-invariant control mode adds a second discrete mode state, $m_{T}$, for each mode of the original dynamical system.}
\label{fig:hybridcontroller}
\end{figure}
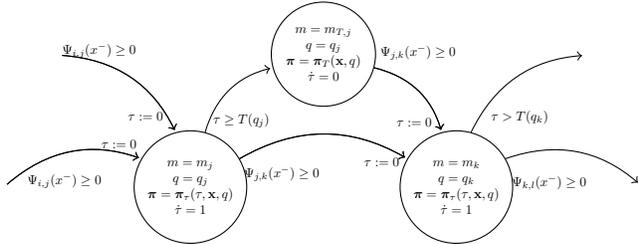

\tikzstyle{block} = [draw, fill=blue!20, rectangle, minimum height=18em, minimum width=25em]
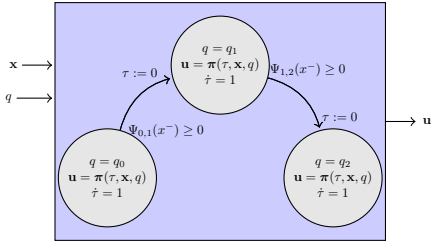
\begin{figure}[t]
\centering
\begin{tikzpicture}[shorten >=1pt,node distance=1.5cm,on grid,auto, scale = 0.5, every node/.style={scale=0.5}] 
   \node [block, align=center,yshift=1.5cm] (controller) {};
   \node[align=center] (center0)  {}; 
   \node[state,fill=gray!20] (q_0) [align=center, left=of center0,label={[align=center]}] {$q=q_0$\\$\bu=\bpi(\tau, \bx,q)$\\$\dot{\tau}=1$}; 
   \node[state,fill=gray!20] (q_1) [align=center, above=of center0,label={[align=center]}] {$q=q_1$\\$\bu=\bpi(\tau,\bx,q)$\\$\dot{\tau}=1$}; 
   \node[state,fill=gray!20] (q_2) [align=center, right=of center0,label={[align=center]}] {$q=q_2$\\$\bu=\bpi(\tau,\bx,q)$\\$\dot{\tau}=1$};    
   
   \node[] (input1) [align=center, left=of q_1,label={[align=center]},xshift=-2.5cm] {$\bx$};
   \node[] (input2) [align=center,above left=of q_0,label={[align=center]},xshift=-0.5cm] {$q$};
   \node[] (output1) [align=center, right=of controller,label={[align=center]},xshift=2.5cm] {$\bu$};
   \draw [->] (input1) --  (input1-|controller.west);
   \draw [->] (input2) --  (input2-|controller.west);
   \draw [->] (controller.east) --  (output1);
    \path[->] 
    (q_0) edge  [align=center, bend left, below,pos=0.15] node {$~~~~~~~~~~~~~~~~~~\Psi_{0,1}(x^{-})\geq 0$} (q_1)
    (q_0) edge  [align=center, bend left, above,pos=0.85] node {$\tau:=0~~~~~~~~~$} (q_1)
	(q_1) edge  [align=center, bend left, above,pos=0.15] node {$~~~~~~~~~~~~\Psi_{1,2}(x^{-})\geq 0$} (q_2)
    (q_1) edge  [align=center, bend left, above,pos=0.95] node {$~~~~~~~~~\tau:=0$} (q_2);

\end{tikzpicture}
\label{fig:hybridcontrollerwaterexit}
\caption{Diagram of the hybrid control system for the water exit maneuver with mode schedule $\xi=\{0,1,2\}$.}
\end{figure}

\begin{align}
\bpi(\tau, \bx,q) =
\begin{cases} 
\bpi_{\tau}(\tau,\bx,q) & \tau < T(q)\\
\bpi_{T}(\tau,\bx,q) & \tau\geq T(q)\\
\end{cases} 
\end{align}
where 
\begin{align}
\dot{\tau} =
\begin{cases} 
1 & \tau < T(q)\\
0 & \tau\geq T(q)\\
\end{cases} 
\end{align}
and
\begin{align}
\bpi_{\tau}(\tau,\bx) &= \matr{K}_q(\bx-\bx_{0}(\tau,q))+\bu_{0}(\tau,q).
\end{align}
$\matr{K}(\tau,q)$ is found by integrating
\begin{align}
-\dot{\matr{S}}(\tau,q) = \matr{A}(\tau,q)^T\matr{S}(\tau,q)+\matr{S}(\tau,q)\matr{A}(\tau,q)\ldots\nonumber\\-\matr{S}(\tau,q)\matr{B}(\tau,q)\matr{R}(q)^{-1}\matr{B}(\tau,q)^T\matr{S}(\tau,q)+\matr{Q}(q)
\end{align}
backwards in time from $\tau=T(q)$ to $\tau=0$. Here $\matr{A}(\tau,q)=\frac{\partial \bff(\bx_0(\tau,q),\bu(\tau,q))}{\partial \bx}$ and $\matr{B}(\tau,q)=\frac{\partial \bff(\bx_0(\tau,q),\bu(\tau,q))}{\partial \bx}$ and 
\begin{align}
\matr{K}(\tau,q) = \matr{R}(q)^{-1}\matr{B}(\tau,q)^T\matr{S}(\tau,q).
\end{align}
The reset map is given as
\begin{align}
\tau^{+}&= 0.
\end{align}

To compute $\bpi_{T}(\tau,\bx,q)$ we only consider the reduced state space $\vect{x}_p = \{\phi, \theta, \psi, \delta_{1}, \delta_{2}, v_{x}, v_{y}, v_{z}, \omega_x, \omega_y, \omega_z\}$. We then use the orientation and velocities given at the guard by $\bx_0(T(q),q)$ to calculate a trim condition. This trim condition becomes the goal state for our time-invariant controller. To compute the gains for the time-invariant controller, we linearize our system about the trim condition and utilize a standard LQR formulation. 

This time-invariant controller should provide additional robustness for the feedback control design. Modeling errors may prevent the time-varying controller from successfully driving the system to a mode transition in finite time. In this case, so long as the time-varying controller is able to drive the system to the region-of-attraction of the time-invariant controller, the time-invariant controller will be able to ensure that the mode transition occurs.

\begin{figure}
\centering
\begin{minipage}{.5\textwidth}
  \centering
  \includegraphics[width=3.25in,trim={30mm 40mm 30mm 70mm},clip]{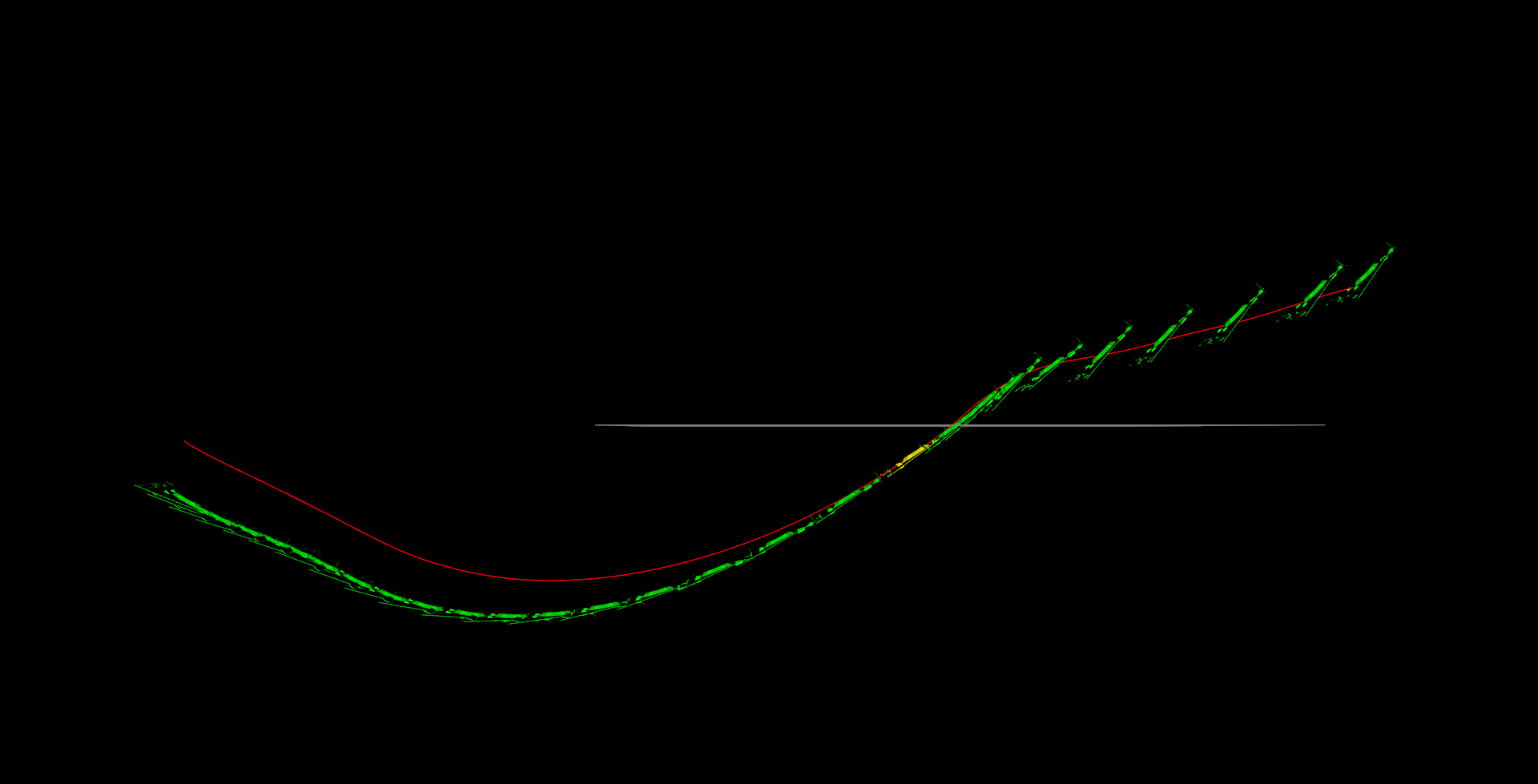}
\end{minipage}%
\vspace{2mm}
\begin{minipage}{.5\textwidth}
  \centering
   \includegraphics[width=3.25in,trim={30mm 40mm 30mm 70mm},clip]{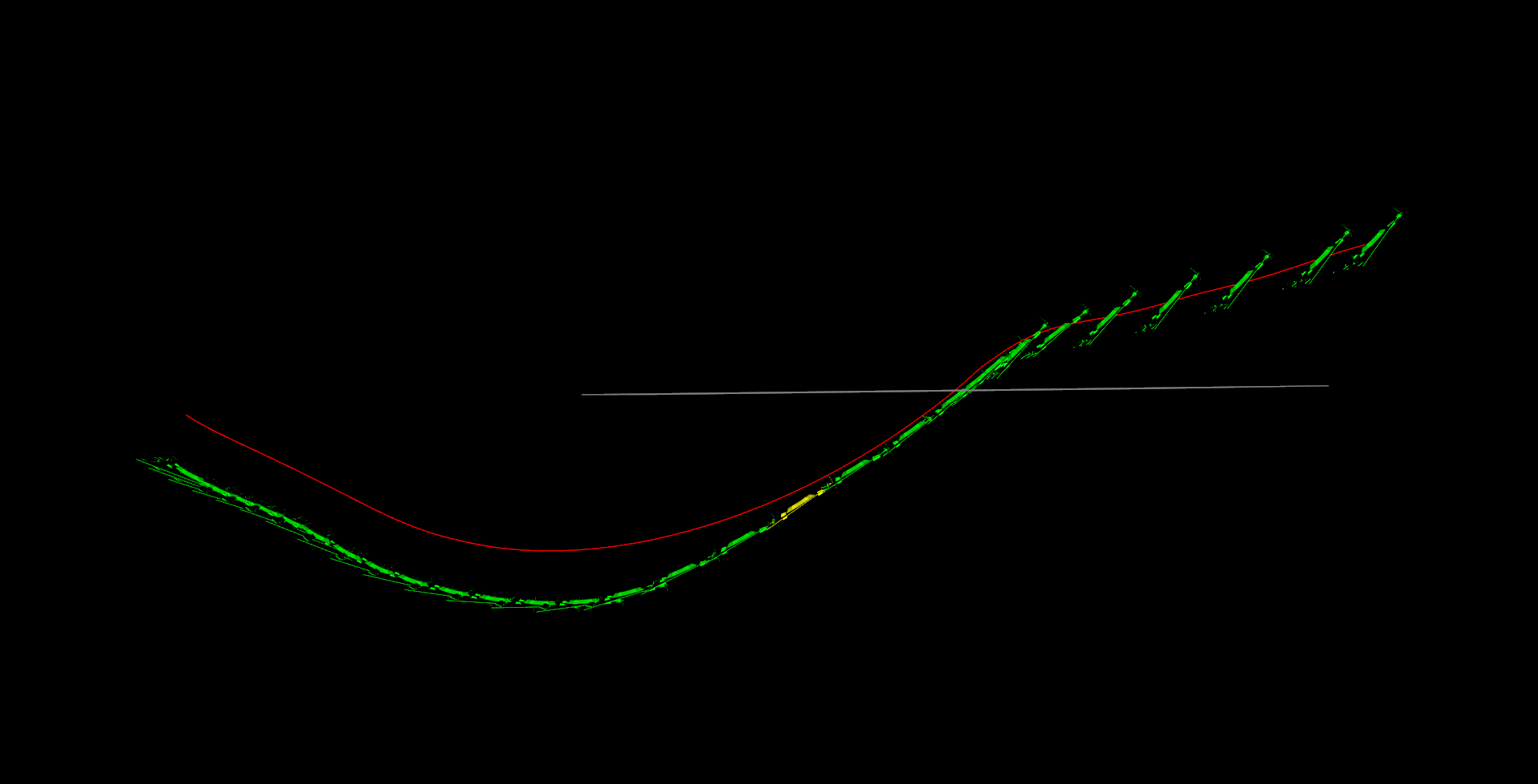}
\end{minipage}
  \caption{Simulation results of our hybrid trajectory-following controller. The red-line represents the nominal trajectory and the vehicle pose highlighted in yellow represents the point at which the TVLQR controller switches to the intermediate time-invariant control mode to ensure domain transition. The gray line represents the separation between air and water. The top and bottom image show results for a system simulated with increasing amounts of additional vehicle drag to demonstrate the importance of the time-invariant intermediate control mode.}
  \label{fig:feedback_control1}
\end{figure}

In Figure \ref{fig:feedback_control1} we show the importance of the time-invariant controller to varying model parameters. In both figures, the termination of the first time-varying control mode is indicated by the yellow vehicle pose. Both figures show the time-varying control law terminating before the propeller breaks the surface due to an increase in vehicle skin friction drag. The time-invariant underwater control mode, however, is able to ensure that the mode-transition from water-to-air still occurs. 

%

\section{STATE ESTIMATION APPROACH}
\begin{figure}
\centering
 \includegraphics[width=2.85in,trim={0mm 10mm 0mm 10mm},clip]{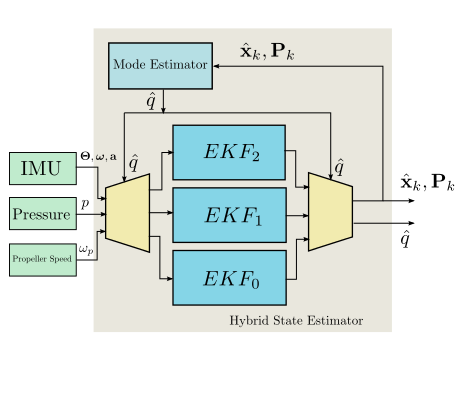}
  \caption{A diagram of the hybrid state estimator. Three EKF filters with different measurement models are used along with a mode estimator to determine the continuous state and discrete mode of the system.}
  \label{fig:estimator_block_diagram}
\end{figure}

The hybrid state estimator  consists of an extended Kalman filter for each mode vehicle mode and a mode estimator that determines the ``active'' filter (see Figure \ref{fig:estimator_block_diagram}).

\subsection{Process Model}
 We used the same process model for each estimator, given as
\begin{align}
\dot{\bx}= \begin{bmatrix}\matr{I} &\matr{0}\\ \matr{0} & \matr{0} \end{bmatrix}+\begin{bmatrix}\matr{0}\\ \matr{R}_b^r(\vectg{\theta}) \end{bmatrix}\vect{a}.
\end{align}
Here $\vect{a}$ represents the linear body accelerations as measured by an inertial measurement unit (IMU). The orientation $\vectg{\theta}$ measured by the IMU is also modeled as an input to the filter. 

\subsection{Measurement Model}
The table below (\ref{Table:sensors}) shows which sensors we assume to be available for each mode.
\begin{center}
 \begin{tabular}{|| c | c | c | c | c | c | c | c ||} 
 \hline
 Mode & $r_x$ & $r_y$ & $r_z$ & $\vectg{\theta}$ & $\vect{v}$ & $\vectg{\omega}$ & $\vect{a}$  \\ [0.5ex] 
  \hline
  water & No & No & Yes & Yes & Yes & Yes & Yes \\ 
 \hline
  water-to-air & No & No & Yes & Yes & No & Yes & Yes \\
 \hline
  air & Yes & Yes & Yes & Yes & No & Yes & Yes \\
  \hline
\end{tabular}
\label{Table:sensors}
\end{center}
We also assume that mode sensors exist which are capable of detecting a zero cross of each guard function  $\Psi_1$ and $\Psi_2$.
Our proposed measurement models are:
\begin{align}
\vect{y}_{q0} = \begin{bmatrix}\rho_\water g z_r\\ {\matr{R}_b^r}^T\dot{\vect{r}}\end{bmatrix}\quad 
\vect{y}_{q1} =\begin{bmatrix}\rho_\water g z_r\\ R_{b,33}^r z_r\end{bmatrix}\quad
\vect{y}_{q2} = \begin{bmatrix}z_r\\y_r\\ R_{b,33}^{rT} z_r\end{bmatrix}
\end{align}
We make the assumption that lateral and longitudinal GPS measurements are available out of the water and that the body velocity is measurable in the water. However, in mode $q_1$, there are no measurements of lateral or longitudinal velocity. Therefore, in our model, we assume that lateral motion is negligible and that the velocity acts along $\bex$. 

To select between modes, we use the standard deviation provided by the variance of the states to determine whether or not a mode transition is likely. This information is fused with data from the mode transition sensors to provide a more robust means of determining whether or not a mode transition has occurred; a mode transition is only considered to have occurred if the guard function is within $2\sigma$ of the zero-crossing based on the state estimates and the mode transition sensors confirm that a transition has occurred.

\section{HARDWARE EXPERIMENTS}
To test our closed-loop control approach, we utilized a prototype similar to what was presented in \cite{moore2018design}. This delta-wing design has an aspect ratio (AR) of 2.4, a wing span of 0.61 m and a center chord length of 0.5 m. The vehicle was constructed using a carbon fiber-foam-carbon fiber sandwich structure. A rigid vertical stabilizer extends above and below the wing and was constructed from a multi-layer carbon-fiber sheet. To control the vehicle's elevons, we use Hitec HS-5065MG servos. The vehicle is propelled by a 6x4 APC propeller with a 4-inch pitch. To drive the propeller, we use a T-Motor MN1806 and a HobbyWing FlyFun 18A speed controller.
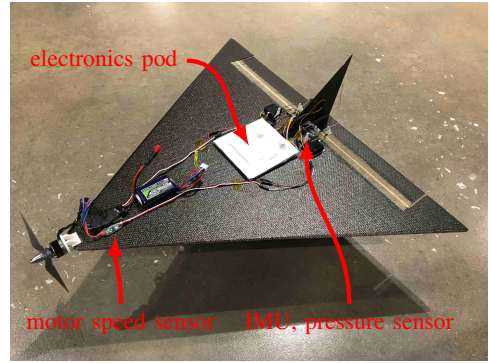
\begin{figure}
\begin{tikzpicture}
\centering
\begin{scope}[]
    \node[anchor=south west,inner sep=0] (image) at (0,0) {\includegraphics[width=2.5in]{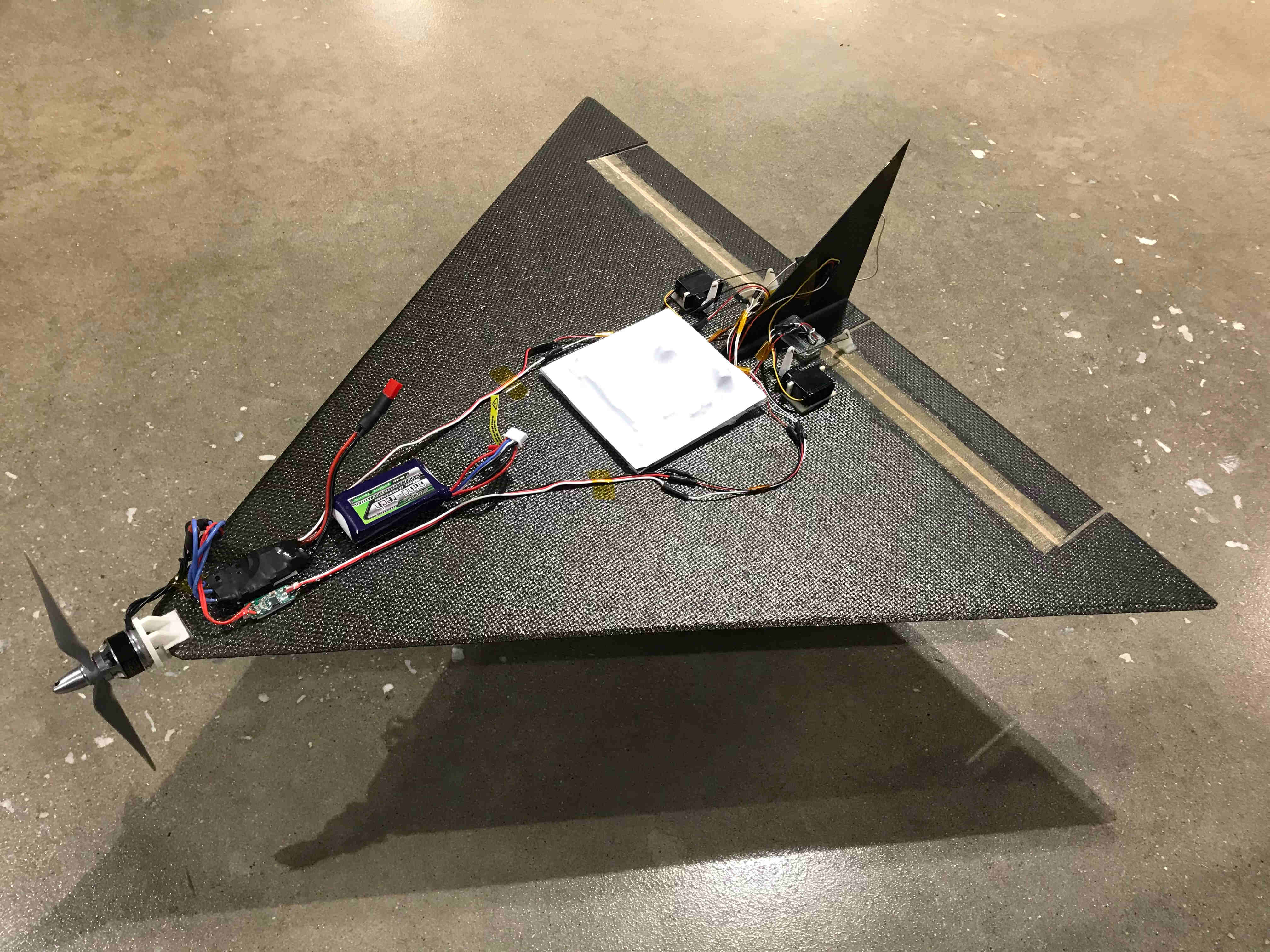}};
\end{scope}
\node [anchor=west,red] (dummy) at (-1,4) {};
\node [anchor=west,red] (pod) at (.15,4) {\small electronics pod};
\node [anchor=west,red] (rpm) at (.1,.5) {\small motor speed sensor};
\node [anchor=west,red] (pressure) at (3.0,.5) {\small IMU, pressure sensor};
    \begin{scope}[x={(image.south east)},y={(image.north west)}]
        \draw [-latex, ultra thick, red] (pod) to[out=0, in=120] (0.5,0.6);
        \draw [-latex, ultra thick, red] (rpm.north) to[out=90, in=-90] (0.22,0.35);
        \draw [-latex, ultra thick, red] (pressure.north) to[out=90, in=-90] (.62,0.62);
    \end{scope}
\end{tikzpicture}
  \caption{Photo of the instrumented UAAV prototype. Figure shows the main electronics pod, motor speed sensor, IMU, and pressure sensor.}
  \label{fig:prototype}
\end{figure}
\subsection{Instrumentation}
To instrument our vehicle, we used a HobbyWing RPM sensor along with the Bosch BNO055 IMU and the TE MS5837-30BA pressure sensor. We built a custom autopilot board consisting of an ATMEGA32U4 for sensing and actuation and a Gumstix Overo Cortex-A8 for executing all the control and state estimation on-board. The fully instrumented system can be seen in Figure \ref{fig:prototype}. To waterproof the electronics, a waterproof pod was constructed by vaccuum forming a thin plastic shell over the electronics. The overall weight of the instrumented vehicle was 375 grams.
\subsection{Propeller Modeling}
\begin{figure}
\centering
\begin{minipage}{.45\textwidth}
  \centering
  \includegraphics[width=3.25in,trim={0mm 0mm 0mm 0mm},clip]{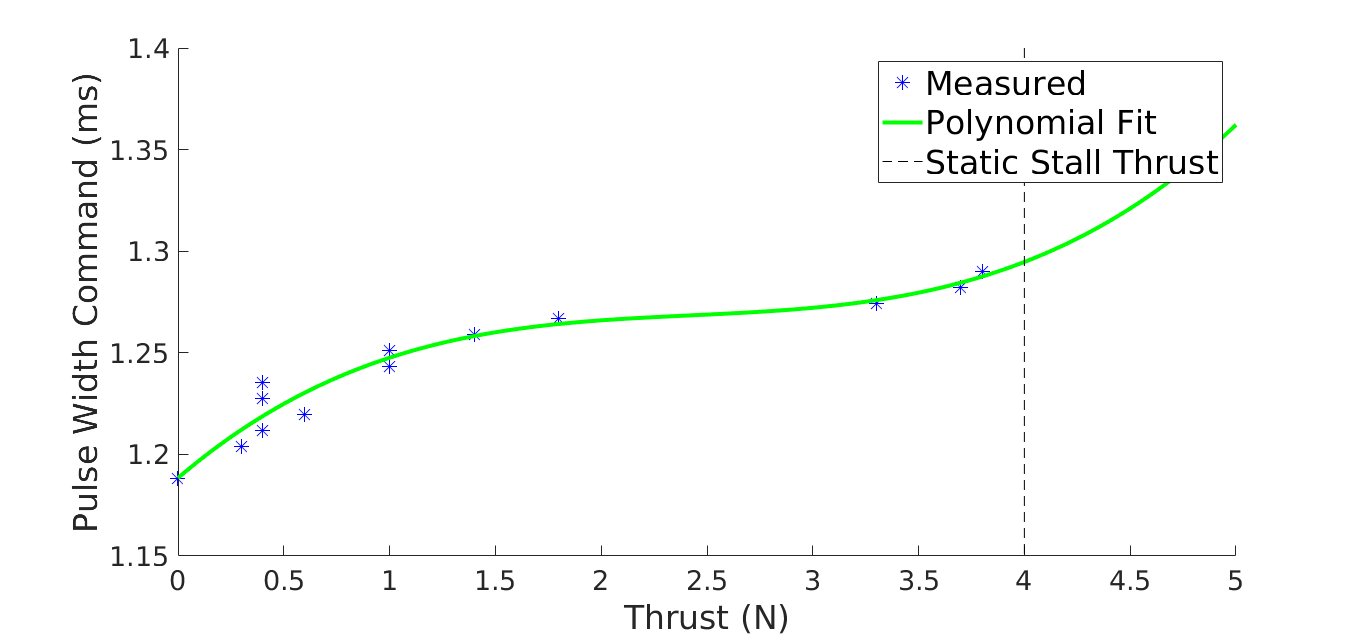}
  \caption{A polynomial mapping from motor controller PWM command to static propeller thrust.}
  \label{fig:propeller_thrust_vs_cmd}
\end{minipage}%
\hspace{2mm}
\begin{minipage}{.45\textwidth}
  \centering
   \includegraphics[width=3.25in,trim={0mm 0mm 0mm 0mm},clip]{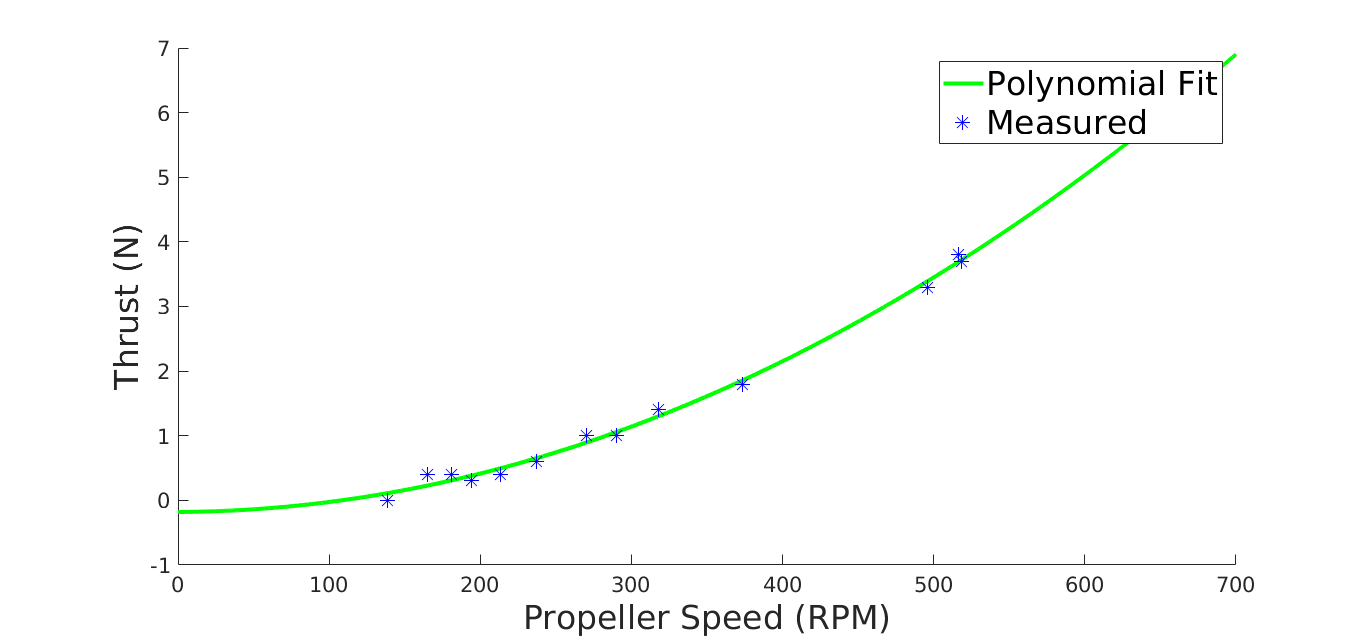}
  \caption{A quadratic polynomial mapping from propeller speed to static propeller thrust}
  \label{fig:propeller_thrust_vs_speed}
\end{minipage}
\end{figure}

To generate a model for the propeller/electric motor propulsion system underwater, a third-order polynomial curve was fit to data and provided a mapping from servo command to thrust (see Figure \ref{fig:propeller_thrust_vs_cmd}).

\subsection{Domain Sensors}
\begin{figure}
\centering
 \includegraphics[width=2.85in,trim={0mm 0mm 0mm 0mm},clip]{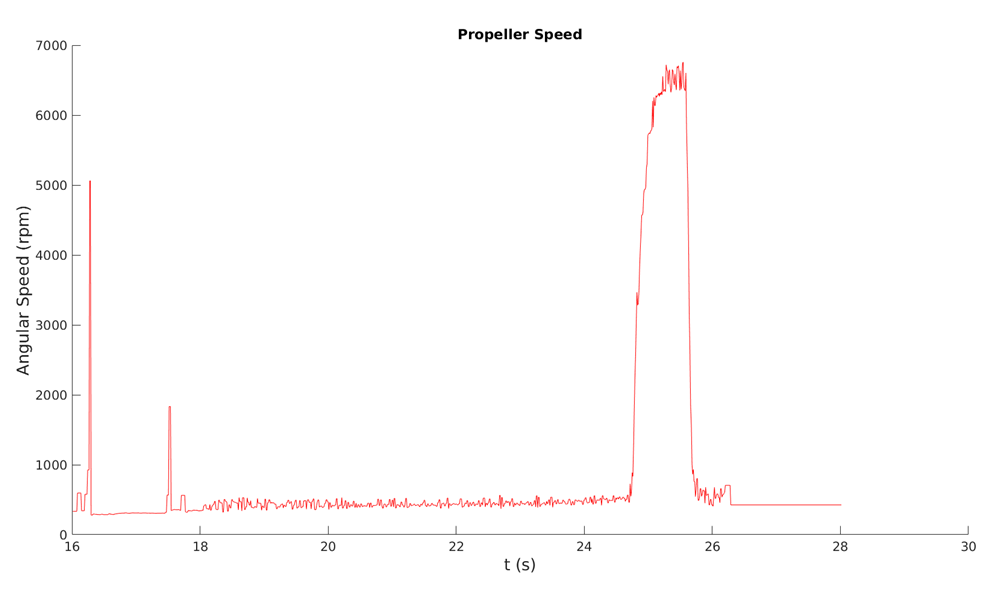}
  \caption{Propeller speed change as motor command is held fixed and propeller exits the water. At about t=25s, the propeller leaves the water and the propeller speed passively jumps to approximated 14.5 times its underwater value.}
  \label{fig:prop_speed}
\end{figure}

To sense the transition from $q_0$ to $q_1$ it was found that the motor itself could serve as a suitable domain sensor. Because the motor runs open loop, when the torque on the propeller changes between fluid mediums, the speed of the propeller increases by $\approx$14x (see Figure \ref{fig:prop_speed}). This drastic change can be easy sensed and used to determine that the propeller has exited the water. While not providing as large a signal-to-noise ratio, the barometric pressure sensor can also be used to sense when the rear of the vehicle exits the water ($q_1$ to $q_2$).

\subsection{Underwater Velocity Measurement}
The propeller also provides a reasonable estimate of the forward speed of the vehicle underwater. If we assume that the UAAV swims with a small angle-of-attack underwater (which holds for our trajectory), and therefore all of the velocity is directed along the body x-axis, then it is possible to measure forward velocity as $\vect{v}=v_x\bex=p\omega_p$, where $p$ is the pitch of the propeller in meters. Figure \ref{fig:horz_vel_est} shows good correlation between the differentiated depth sensor data and the vertical velocity estimated from propeller speed.
\subsection{State Estimation Experiments}
\begin{figure}
\centering
 \includegraphics[width=3.5in,trim={0mm 0mm 0mm 0mm},clip]{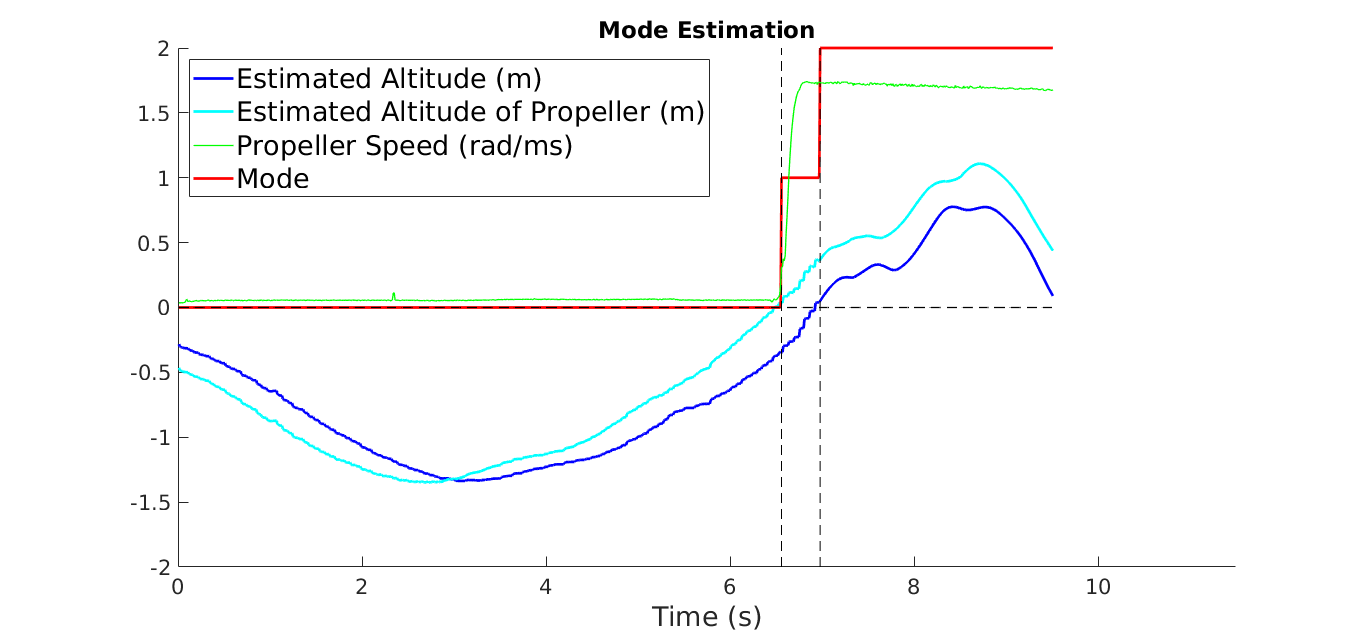}
  \caption{Demonstration of successful mode tracking. The figure shows that when the estimated propeller position crosses the air-water boundary, the propeller increase in speed can successfully be used to detect the mode change. Approximately 0.25s is required to reach maximum propeller speed in air when starting from in water.}
  \label{fig:mode_est}
\end{figure}

Using the aforementioned instrumentation, state-estimation was able to be successfully executed on-board the vehicle. Figure \ref{fig:mode_est} demonstrates effective mode tracking, where the standard deviation in the propeller estimates is used along with the propeller speed change and the barometric pressure sensor to determine the vehicle mode. 

\begin{figure}
\centering
 \includegraphics[width=3.25in,trim={0mm 10mm 0mm 10mm},clip]{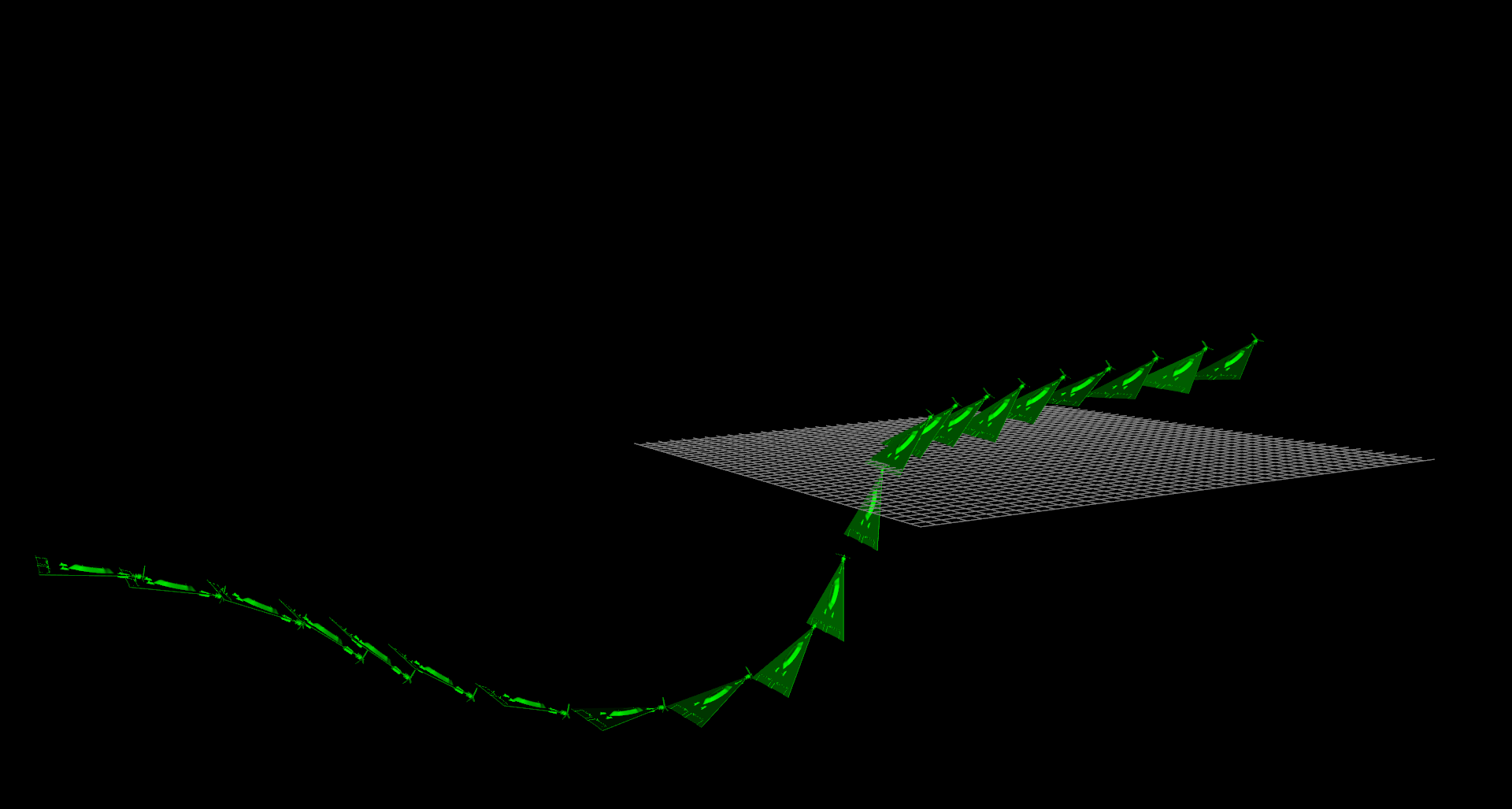}
  \caption{Visualization of the unmanned aerial-aquatic vehicle's estimated pose as the vehicle executes a water-to-air transition. In the flight phase, position and velocity estimates are strictly based on dead-reckoning via the IMU.}
  \label{fig:state_estimation_results}
\end{figure}

\begin{figure}
\centering
 \includegraphics[width=3.5in,trim={0mm 0mm 0mm 0mm},clip]{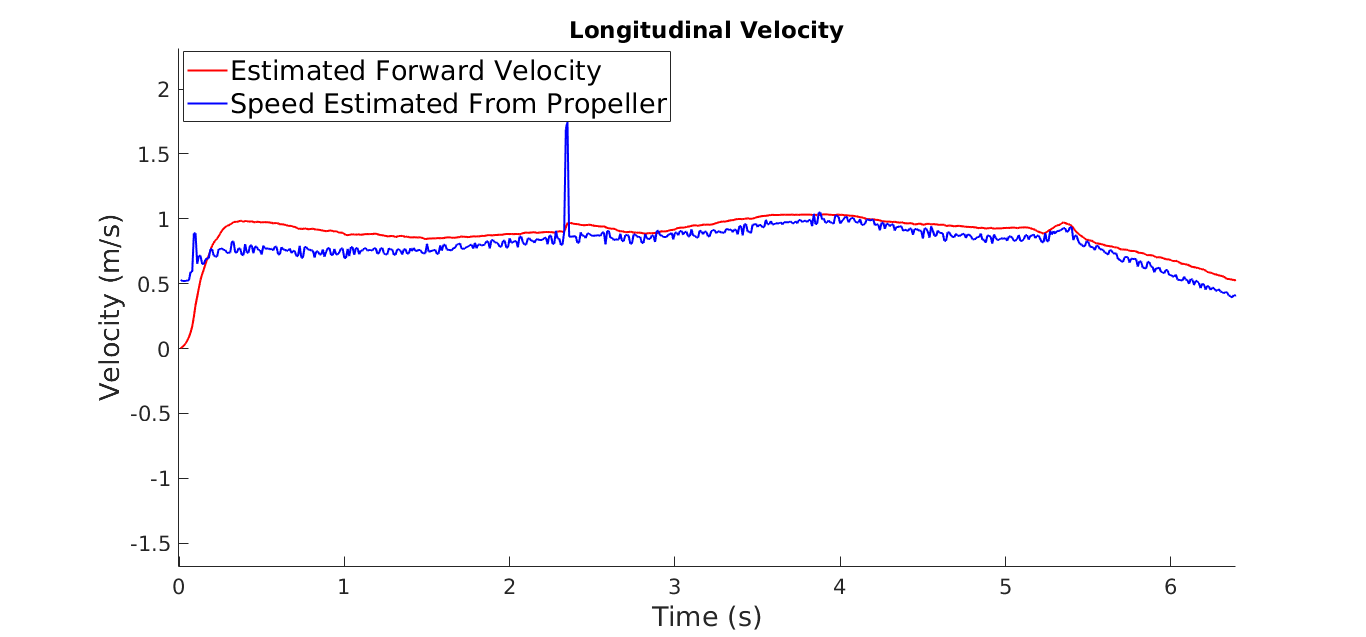}
  \caption{Estimate of the longitudinal velocity of the vehicle compared with the speed estimated from the propeller in the water domain. This demonstrates the ability of the estimator to reject noise from propeller speed sensor.}
  \label{fig:horz_vel_est}
\end{figure}

\begin{figure}
\centering
 \includegraphics[width=3.5in,trim={0mm 0mm 0mm 0mm},clip]{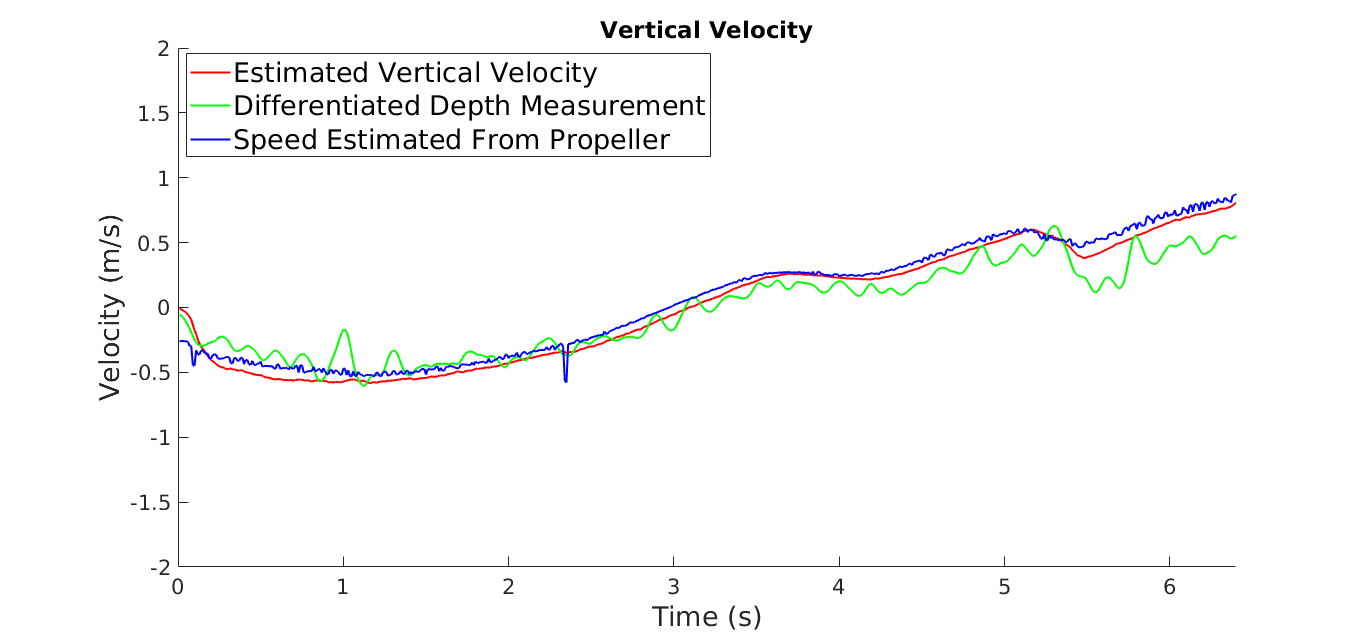}
  \caption{Estimate of the vertical velocity of the vehicle compared to the differentiated depth information and the velocity estimated from the propeller in the water domain.}
  \label{fig:vert_vel_est}
\end{figure}

Figures \ref{fig:horz_vel_est} and \ref{fig:vert_vel_est} show good velocity estimation and demonstrate that the accelerometer biases can be estimated in the water and then used for a short time during the transition from water to air to estimate the vehicle's position (see Figure \ref{fig:state_estimation_results}).


 \begin{figure}
\centering
 \includegraphics[width=3.5in,trim={0mm 0mm 0mm 0mm},clip]{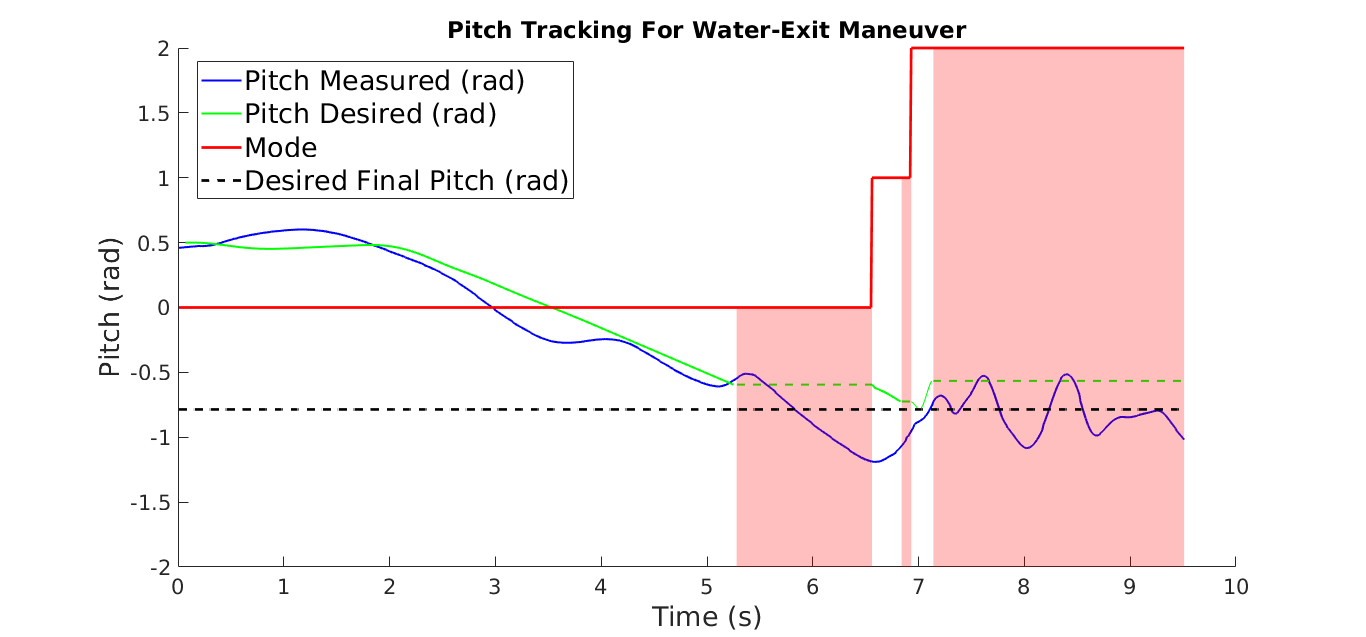}
  \caption{Pitch tracking of vehicle for a water-exit maneuver. Red areas indicate regions where the time-invariant controller takes over.}
  \label{fig:pitch_tracking}
\end{figure}
 \begin{figure}
\centering
 \includegraphics[width=3.5in,trim={0mm 0mm 0mm 0mm},clip]{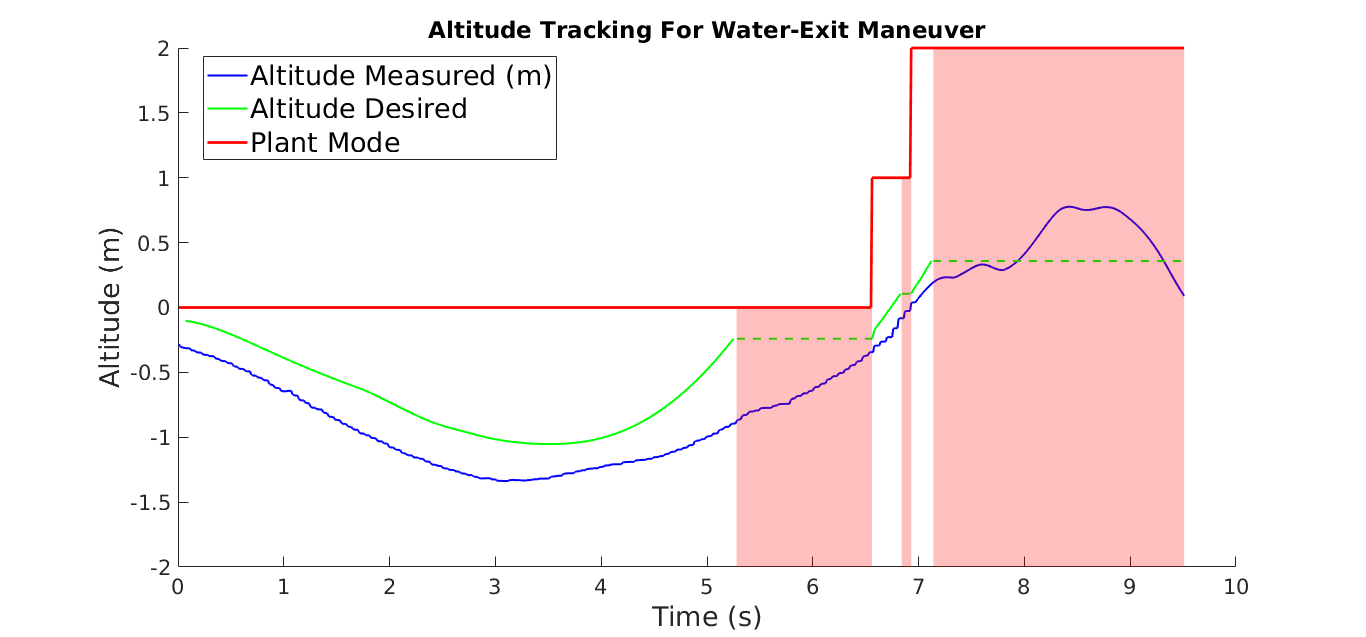}
  \caption{Depth tracking of vehicle for a water-exit maneuver. Red areas indicate regions where the time-invariant controller takes over.}
  \label{fig:altitude_tracking}
\end{figure}

\subsection{Closed-loop Control Experiments}
\begin{figure}
\centering
  \includegraphics[width=3.5in,trim={0cm 0cm 0cm 0cm},clip]{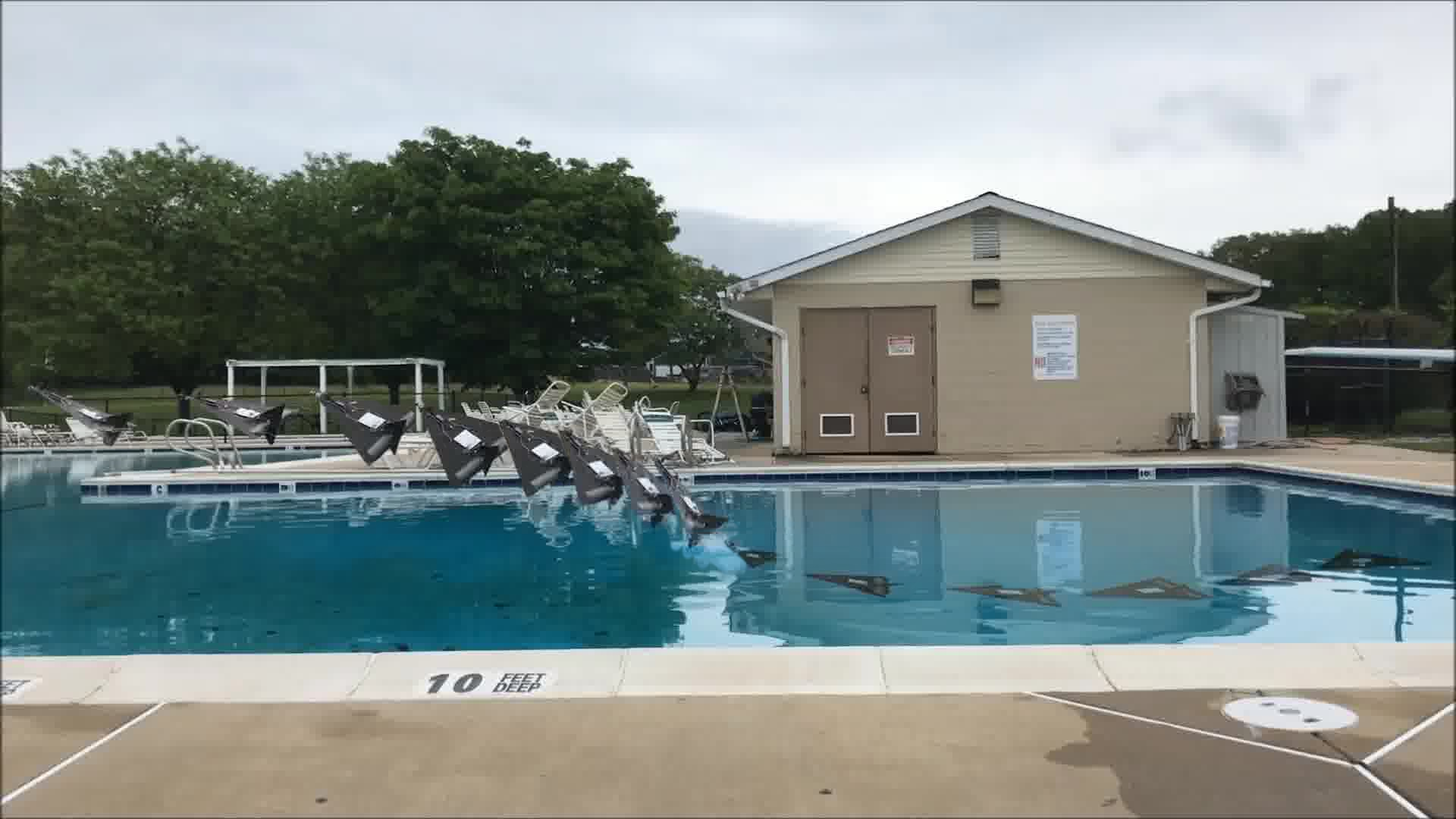}
  \caption{Successful water-exit to 45 degree prop-hang. Time-step between underwater frames is  $\approx$1.67s while the time-step between air frames is $\approx$0.25s}
  \label{fig:timelapse}
\end{figure}

\begin{figure}
\centering
  \includegraphics[width=3.75in,trim={1cm 0cm 1cm 0cm},clip]{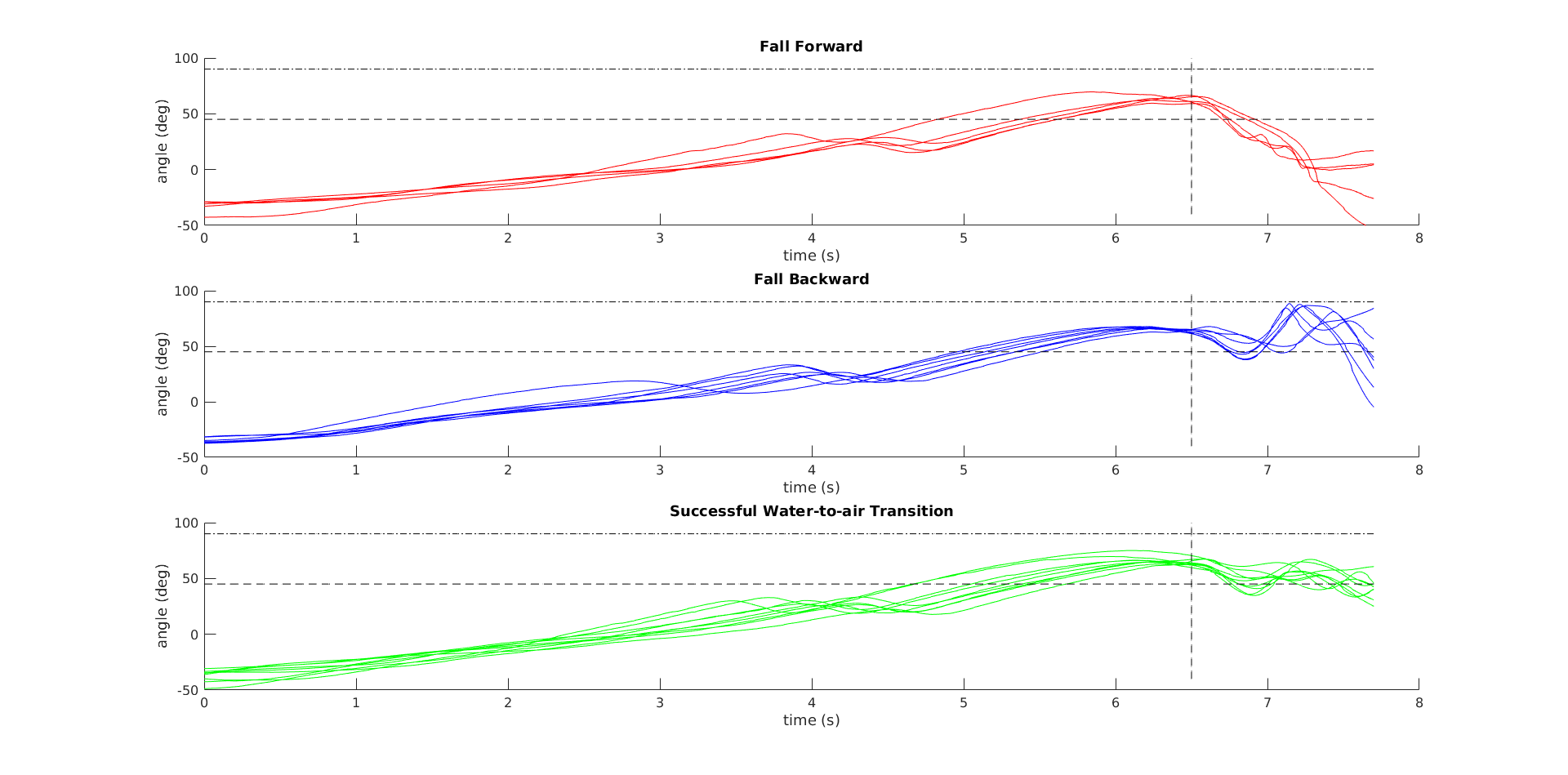}
  \caption{Performance of closed-loop controller over twenty trials. Two primarily failure modes were observed--- falling backwards and falling forwards. Eight of the trials exhibited what would be considered a successful air-to-water transition, where the system entered a flight mode regulating pitch to 45 degrees. The upper dashed line represents 90 degree pitch, the value approached by the vehicle inversion failure cases. The lower dashed line represents the desired 45 degree pitch for flight. The vertical dashed line represents the time at which the vehicle is completely in the air domain.}
  \label{fig:finalresults}
\end{figure}

\begin{figure}
\centering
  \includegraphics[width=3.5in,trim={0cm 0cm 0cm 0cm},clip]{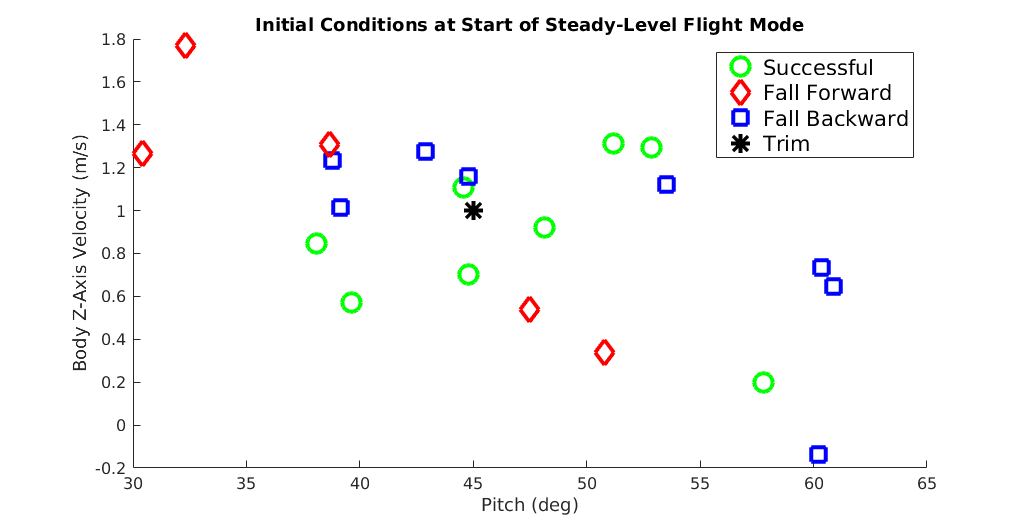}
  \caption{Experimental initial conditions for the final flight controller plotted on a two dimensional slice of the state space along the body-z and pitch axes. Green circles represent initial conditions that successfully transition to steady-level flight. Red diamonds represent initial conditions that fall forward and blue squares represent trials that fall backward. Black represents the trim flight condition at 45 degrees pitch.}
  \label{fig:slice}
\end{figure}

To test our system, we attempted to execute the trajectory shown in Figure \ref{fig:feedback_control1} on the physical hardware. To do this, we held the vehicle at an initial downward pitch, just below the surface of the water. Then, we started the motor and released the vehicle. After the vehicle dove down about one foot, it would start to execute the water-exit maneuver. We were able to execute several successful water autonomous exits. Figures \ref{fig:pitch_tracking} and \ref{fig:altitude_tracking} show the state estimator effectively tracking the system across multiple domains. Figure \ref{fig:timelapse} shows a successful water exit, where the vehicle transitioned to a flight regime where it successfully regulated attitude. Figure \ref{fig:finalresults} shows the results of twenty trial runs, where eight of the trials transitioned to controlled flight (green). Seven of the runs flipped over backward (blue) after exiting the water, and four of the runs fell forward (red), unable to gain enough lift. Figure \ref{fig:slice} shows the state of the vehicle prior to the final flight-mode and labels the initial conditions based on final performance. In this figure, we can see that, while the data is sparse, there does seem to be a clear region-of-attraction around the trim condition of 45 degrees. From inspection, vehicle pitch and body z-axis velocity (which is highly correlated with lift force at high angles of attack), seemed to have the greatest affect on future vehicle performance. 


\section{DISCUSSION}
A number of important observations can be made from inspecting the experimental results. First, as is evident from Figures \ref{fig:pitch_tracking} and \ref{fig:altitude_tracking}, it is common for the time-varying ``underwater'' trajectory to complete early (i.e. before the propeller breaks the surface). This was also the case in our simulated example (see Figure \ref{fig:feedback_control1}). Without the time-invariant control modes to ensure the completion of the underwater mode, domain transition would not occur. 

Second, we also observed large transients during the hand-off between the time-varying and time-invariant controllers (see Figure \ref{fig:altitude_tracking}). This was most likely due to the relatively large cost on pitch required for the time-invariant LQR controller to ensure consistent pitch at water-exit. To improve this, a more accurate model of the vehicle could be obtained through better system identification. This more accurate model would then reduce the temporal discrepancy at the termination of the time-varying trajectory. Alternatively, an approach such as transverse-linearization \cite{manchester2011transverse} could be employed to parameterize the reference trajectory using state rather than time. It is our belief that utilizing transverse linearization would greatly improve the overall robustness of the feedback control law to modeling errors and hybrid mode transitions. 

Third, we believe a strong case exists for optimizing for robustness when generating the reference trajectory, as explored in \cite{manchester2017dirtrel}. For instance, the closer the vehicle is to 45 degrees pitch when it exits the water (see Figure \ref{fig:slice}), the more likely it is to transition into a prop-hang and then forward flight. However, due to unmodeled dynamics in the transition, there is often insufficient clearance between the trailing edge of the wing and the water. This can lead to ``reentry'' of the trailing edge into the  water and a ``fall-forward'' failure mode. A higher pitch at exit ($\approx$ 70 degrees) allows for better utilization of the thrust during the transition and can provide more clearance between the water and the wing's trailing edge prior to the prop-hang mode (see the ``success'' outlier in Figure \ref{fig:slice}).

\section{CONCLUSION}

In this paper, we have designed and demonstrated a control system for a delta-wing unmanned aerial-aquatic vehicle. We have shown that it is possible to use feedback control and a simple vehicle design to achieve dynamic transition between water and air with entirely onboard sensing. In future work, we plan to apply transverse-linearization and robust trajectory design to our system. We also plan on investigating a number of hardware design improvements. For instance, a second rear-propeller could help by maintaining thrust underwater during water-exit. An ultrasound sensor (as demonstrated in \cite{ore2015autonomous}) could also provide better state estimation by providing height above the water during exit and flight. With some of these improvements, we believe that we will not only be able to improve the reliability of our water-exit, but that we will be able to achieve multiple domain transitions in the presence of significant disturbances, such as wind and waves.

\addtolength{\textheight}{-12cm}

\section*{ACKNOWLEDGMENT}
We would like to thank Gary Persing, the Eastwood Community Association, and the South Carroll Swim Club for the use of their pools. We would also like to thank Kevin Wolfe for his ideas and insights into vacuum forming for electronics packaging and Craig Leese for his help with vehicle fabrication. This project was funded by JHU/APL internal research.
\bibliographystyle{IEEEtran}
\bibliography{main}
\nopagebreak
\end{document}